\documentclass[conference]{IEEEtran}
\IEEEoverridecommandlockouts
\usepackage{cite}
\usepackage{amsmath,amssymb,amsfonts}
\usepackage{algorithm}
\usepackage{algorithmic}
\usepackage{graphicx,pifont}
\usepackage{textcomp}
\usepackage{xcolor}
\usepackage{makecell}
\usepackage{multicol}
\usepackage{multirow}
\usepackage[hyphens]{url}
\usepackage{color,soul}

\usepackage[a4paper, total={184mm,239mm}]{geometry}
\def\BibTeX{{\rm B\kern-.05em{\sc i\kern-.025em b}\kern-.08em
    T\kern-.1667em\lower.7ex\hbox{E}\kern-.125emX}}
\begin{document}

\title{DeepCAM: A Fully CAM-based Inference Accelerator with Variable Hash Lengths for Energy-efficient Deep Neural Networks
\vspace{ -3mm}}
\author{
Duy-Thanh Nguyen, Abhiroop Bhattacharjee, Abhishek Moitra, and Priyadarshini Panda \\
\{duy-thanh.nguyen, abhiroop.bhattacharjee, abhishek.moitra, priya.panda\}@yale.edu \\

Department of Electrical Engineering, Yale University, USA
\vspace{ -3mm}
}
\maketitle


\begin{abstract}

With ever increasing depth and width in deep neural networks to achieve state-of-the-art performance, deep learning computation has significantly grown, and dot-products remain dominant in overall computation time. Most prior works are built on conventional dot-product where weighted input summation is used to represent the neuron operation. However, another implementation of dot-product based on the notion of angles and magnitudes in the Euclidean space has attracted limited attention. This paper proposes {\em DeepCAM}, an inference accelerator built on two critical innovations to alleviate the computation time bottleneck of convolutional neural networks. The first innovation is an approximate dot-product built on computations in the Euclidean space that can replace addition and multiplication with simple bit-wise operations. The second innovation is a dynamic size content addressable memory-based (CAM-based) accelerator to perform bit-wise operations and accelerate the CNNs with a lower computation time. Our experiments on benchmark image recognition datasets demonstrate that DeepCAM is up to 523$\times$ and 3498$\times$ faster than Eyeriss and traditional CPUs like Intel Skylake, respectively. Furthermore, the energy consumed by our DeepCAM approach is 2.16$\times$ to 109$\times$ less compared to Eyeriss.

\end{abstract}
\section{Introduction}
Deep learning has surpassed humans in various domains, such as image classification, natural language processing, and data generation~\cite{lecun2015deep}. However, this phenomenal progress has also led to a significant increase in the parameter size of a deep neural network (DNN) model in terms of its depth (layers) and width (filters). Dot-product computations in DNNs are highly computation-intensive accounting for more than 90\% of the time to process various DNN workloads~\cite{krizhevsky2012imagenet}. There have been various hardware accelerators for DNN inference such as Eyeriss~\cite{chen2016eyeriss}, TPU~\cite{jouppi2021ten}, Rapid~\cite{venkataramani2021rapid} among others to reduce the dot-product computation time in large-scale DNN deployment. However, conventional von-Neumann inference accelerators incur significantly high memory access energy. Specifically in such architectures, the on-chip memory (SRAM) and off-chip (DRAM) accesses incur 6$\times$ and 200$\times$ higher energy consumption compared to dot-product operation \cite{chen2016eyeriss}. 


Typical systolic array-based accelerators with N$\times$N processing-arrays 
can achieve a computational time of O(N) to carry out dot-product operations. To this end, designing an architecture to further reduce the dot-product computational time to O(1) in traditional von-Neumann architectures with higher energy-efficiency has been a challenge for researchers. Recently, Kai Ni et al.~\cite{ni2019ferroelectric} have proposed a sense amplifier for time sensing using a content addressable memory (CAM) based architecture to estimate the hamming distance between a search key and the CAM-data with high parallelism. The work by Kai Ni et al. ~\cite{ni2019ferroelectric} opens up doors for us to achieve O(1) computation time for dot-products with high parallelism. In this regard, we look into a different kind of dot-product implementation, called geometric dot-product. Typically, all DNN systolic array accelerators are designed to implement algebraic dot-products \cite{reuther2020survey}, that essentially involves multiply-and-accumulate (MAC) operations. We show that for achieving dot-product computation time of O(1) with a CAM-based architecture, the geometric implementation comes handy. In the geometric dot-product, operands are treated as vectors with magnitudes and directions. The dot-product of two operands (vectors) can, thus, be computed using their magnitudes and the angle between them. Based on this definition, the angle between two vectors can be estimated using our CAM-based architecture. This crucial concept allows us to achieve significantly higher throughput and better energy-efficiency during DNN inference, compared to state-of-the-art Eyeriss accelerator~\cite{chen2016eyeriss}.

In this paper, we propose DeepCAM, a novel Process-In-Memory (PIM) based inference accelerator architecture using CAMs, to replace standard algebraic dot-product operations with approximate dot-products (based on geometric implementation) to speed up the DNN computation time and reduce the inference energy. We highlight our key contributions as follows:
\begin{itemize}
    \item We propose an approximate implementation of dot-products with variable hash lengths (based on geometric implementation) for CNN inference on DeepCAM, without significant loss in classification accuracy.
    \item We propose a dynamic size CAM-based inference accelerator with re-configurable hash lengths for processing dot-products with O(1) time-complexity.
    
    
    \item We evaluate our DeepCAM accelerator on various CNN architectures- LeNet5, VGG11, VGG16 and ResNet18, using benchmark datasets (MNIST, CIFAR10 and CIFAR100). We obtain $\sim523\times$ lower computation time and $\sim109\times$ better energy-efficiency per inference compared to the state-of-the-art Eyeriss \cite{chen2016eyeriss} accelerator. 
    
    \item We also compare our DeepCAM accelerator against previously proposed analog PIM-based CNN inference accelerators \cite{peng2019dnn+, valavi201964}. For VGG11 CNN inferred with CIFAR10, DeepCAM is $\sim71.68\times$ and $\sim7.27\times$ more energy-efficient than \cite{peng2019dnn+} and  \cite{valavi201964}, respectively. 
    
    
\end{itemize}

The remainder of the paper is organized as follows. Firstly, we briefly discuss the background on CAMs and dot-product operations in section~\ref{sec:background}. Secondly, we explain our DeepCAM-related problem in section~\ref{sec:deepcam}. Thirdly, we provide the evaluation methodology and results in section~\ref{sec:evaluation}. We further discuss the related works and comparison in section~\ref{sec:relatedworks}. Finally, we will conclude our work in section~\ref{sec:conlcusion}.

\section{Background \& Motivation}
\label{sec:background}
\subsection{CAM/TCAM - beyond CMOS and non-CMOS technology}
\label{sec:cambackground}

\begin{figure}[!t]
\centering
\includegraphics[width=1.0\linewidth]{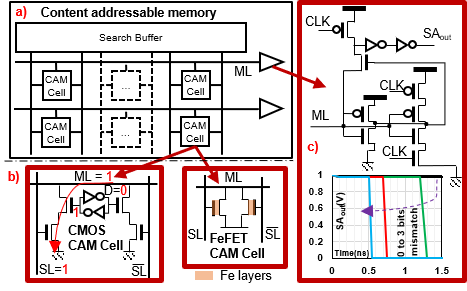}
\caption{a) Block diagram of content addressable memory(CAM) architecture. b) CMOS and FeFET CAM cells. c) Clocked self-referenced sense amplifier(SA) for detecting the matching degrees.}
\vspace{ -3mm}
\label{fig:camcell}
\end{figure}

A content-addressable memory (CAM) shown in Fig \ref{fig:camcell}.a facilitates parallel searching of a query data with content stored in the CAM memory. There are two types of CAM: 1) Binary CAM (or CAM) matching values 0 or 1 only \cite{karam2015emerging}, and 2) Ternary CAM (or TCAM) matching values 0, 1, or X (don't care). In the basic CMOS CAM cell design, the CAM cell includes a storage node (normally SRAM) and a pull-down CMOS circuit as a peripheral. During the search operation, the match line (ML) is first pre-charged to Vdd. The ML remains charged if the query in the search line (SL) and the content in the memory match and discharged otherwise. Due to the parallel search capability, CAMs can achieve O(1) computation time complexity. Besides search operation, CAM can be used to calculate the hamming distance~\cite{ni2019ferroelectric}. Here, the correlation between the number of bit mismatch and the time to pull down the ML voltage is leveraged. Based on this observation, ~\cite{ni2019ferroelectric} proposed the clocked self-referenced sense amplifiers as shown in Fig~\ref{fig:camcell}.c converting the pulled down time to hamming distance.

In the CMOS design, CAM and TCAM require 9-10 and 16 transistors, respectively. Because the CMOS memory cell is usually 2-10$\times$ larger than the non-volatile memory cell \cite{khan2020future}, the CMOS implementation of CAM/TCAM will incur significant overhead. However, the implementation of CAM and TCAM in non-volatile memory technology requires two transistors and two non-volatile memory nodes, as shown in Fig~\ref{fig:camcell}.b. Thus, non-volatile memory nodes are preferred over CMOS transistors in implementing CAM/TCAM. As reported in ~\cite{yin2020fecam}, using FeFETs reduces the cell size to 7.5$\times$ with 2.4$\times$ lesser search energy than CMOS. With the premise of both energy saving and hamming-distance estimation in parallel, it is possible to build up a fast and energy-efficient deep learning accelerator. Thus, we consider FeFET CAM in this paper and the design details are provided in section~\ref{sec:deepcam}.

\subsection{Dot-product and its approximation with random projection}
\label{sec:dotproduct}

As a fundamental operation for convolution and fully connected layers in CNNs, the algebraic dot-product is computed by the MAC operation between input activation and weight vectors. Assuming \textbf{x} and \textbf{y} to be two vectors with N elements, their algebraic dot-product is defined as follows:
\vspace{-2mm}
\begin{equation}
\label{equ:dotproduct1}
    \textbf{x}.\textbf{y} = \sum_{i=1}^{N} x_iy_i
\vspace{-3mm}
\end{equation}

In Euclidean space, vectors are represented with magnitudes and angular components (directions). The magnitude of a vector is its L2 norm ($\lVert . \rVert_2^2$), and the angular component is defined as the cosine of the smallest angle between two vectors. Hence, we define geometric dot-product as follows:
\vspace{-2mm}
\begin{equation}
\label{equ:dotproduct2}
    \textbf{x}.\textbf{y} = \lVert x \rVert_2^2 \lVert y \rVert_2^2 cos(\theta)
    \vspace{-2mm}
\end{equation}

If \textbf{x} and \textbf{y} are replaced with input activations and weights for a DNN model, then finding (or estimating) $cos(\theta)$ for the computation of geometric dot-product is a tedious task. However, this problem can be solved by the Johnson-Lindenstrauss(J-L) lemma~\cite{johnson1984extensions}. For better understanding, let us define the mapping of $\textbf{x} \in R^n$ to $\textbf{Z} \in \{0,1\}^k$ as a hashing method that maps the n-dimensional \textbf{x} vector to a k-dimensional \textbf{Z} vector. Say, the conversion hashing function is a matrix C $\in R^{n \times k}$ and C follows a normal distribution $\sim N(0,1)$. Any \textbf{x} vector can be converted into hyperspace \textbf{Z} by taking the signed bits of the projection product of \textbf{x} and C matrix: $hash(x) = sign(xC)$. Because C is a random matrix, we call this operation as random projection with hash length (k). The angle between two vectors $\theta$ can thus be approximated as the hamming distance (HD) between two hashed vectors \textbf{x} and \textbf{y} \cite{goemans1995improved}:
\vspace{-1mm}
\begin{equation}
\label{equ:appangle}
    \theta_{\textbf{x}.\textbf{y}} = \pi Pr(hash(x) \ne hash(y)) \approx \frac{\pi}{k}HD(hash(x), hash(y))
\end{equation}

\begin{figure}[!t]
\centering
\includegraphics[width=0.9\linewidth]{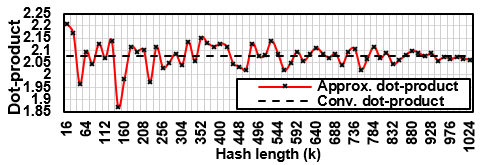}
\vspace{-5mm}
\caption{Plot showing output result comparison between the approximate dot-product and conventional (algebraic) dot-product.}
\label{fig:appdotpro}
\end{figure}

From eq. \ref{equ:dotproduct2} \& \ref{equ:appangle}, we approximate geometric dot-product as follows:
\vspace{-3mm}
\begin{equation}
\label{equ:appdotproduct}
     \textbf{x}.\textbf{y} \approx \lVert x \rVert_2^2 \lVert y \rVert_2^2 cos(\frac{\pi}{k}HD(hash(x), hash(y)))
\end{equation}
Now, let us consider the following example: If \textbf{x} = [0.6012, 0.8383, 0.6859, 0.5712], \textbf{y} = [0.9044, 0.5352, 0.8110, 0.9243], the conventional algebraic dot-product will be 2.0765. We find in Fig.~\ref{fig:appdotpro} that the dot-product approximation using eq. \ref{equ:appdotproduct} is nearly equal to the result of the algebraic dot-product, and longer hash lengths (k) lead to better approximation. Based on this approximate geometric dot-product formulation, we develop the CAM-based accelerator for the CNNs in this paper. Furthermore, owing to the error-tolerant characteristic of deep CNNs, we will see that our model's performance does not degrade drastically due to the approximation.

\section{Overview of DeepCAM architecture}
\label{sec:deepcam}

\begin{figure}[!t]
\centering
\includegraphics[width=\linewidth]{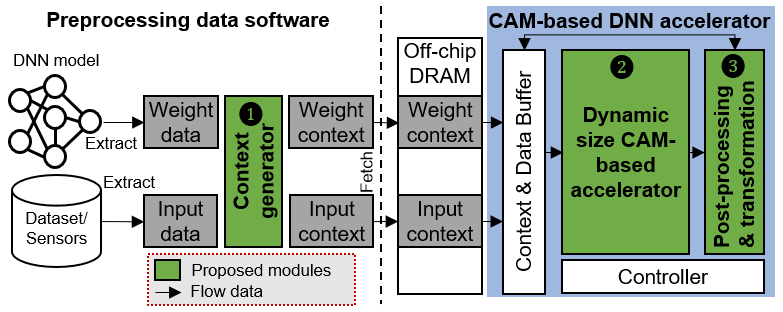}
\vspace{-5mm}
\caption{Full architecture overview of DeepCAM. 1) Context generator software for pre-processing the deep learning data. 2) A dynamic size CAM-based accelerator for dot-product operations. 3) Post-processing and transformation module for post dot-product computations and online activation context generation.}
\label{fig:deepcam}
\vspace{-5mm}
\end{figure}

In this section, we describe our DeepCAM architecture, which is a fully CAM-based PIM accelerator for CNN inference. The design of DeepCAM comprises of three major components: 1) a context generator software, 2) a dynamic sized CAM-based accelerator, and 3) a post-processing and transformation unit. These components are shown in Fig.~\ref{fig:deepcam}. 

\subsection{Context Generator}
\label{sec:contextgen}

\begin{figure}[!t]
\centering
\includegraphics[width=0.8\linewidth]{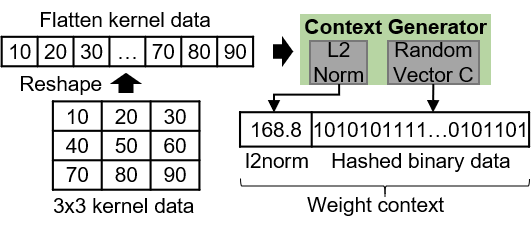}
\caption{An example showing the context generation process.}
\vspace{-5mm}
\label{fig:contextgenerator}
\end{figure}

As discussed in section~\ref{sec:dotproduct}, the approximate dot-product requires the magnitude and hashed binary data for each input activation and weight (see equation~\ref{equ:appdotproduct}). The magnitude is a Euclidean norm or L2 norm with 8-bit minifloat representation~\cite{gysel2018ristretto}. The hashed binary data can be generated by multiplying the activation or weight with a random matrix C.  
As shown in Fig~\ref{fig:contextgenerator}, the context generator is a software that generates the two components: 1) the L2 norms and 2) the hashed binary data, for the input activations and weights. We need to reshape the weight/activation matrices before computing the L2 norm and hashed binary vectors. An example is shown in Fig~\ref{fig:contextgenerator} to describe the process of building a weight context from a kernel of size $3 \times 3$. Note that the contexts for the pre-trained CNN weights and input data can be pre-processed in the software and thus, cause no impact on the computation time during inference on hardware. However, the intermediate activations generated at the end of one CNN layer need to be transformed into the activation contexts before the computation of the subsequent layer. Hence, we propose an online transformation technique (see Post-processing \& transformation unit in Fig.~\ref{fig:deepcam}) for on-the-fly activation context generation, discussed in section~\ref{sec:postprocessing}.

\begin{figure*}[!t]
\centering
\includegraphics[width=\linewidth]{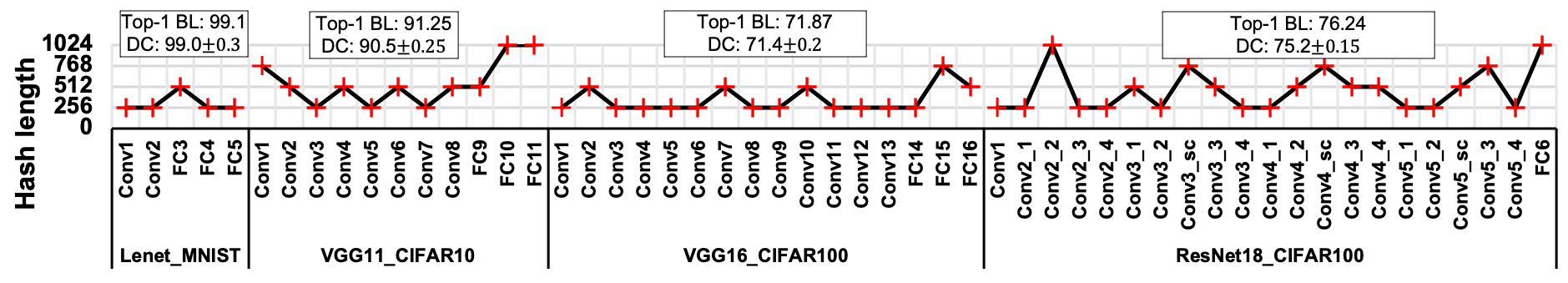}
\caption{Plot showing that variable hash lengths are required to maintain the Top-1 classification accuracy of LeNet5, VGG11, VGG16 and ResNet18 CNN models on DeepCAM. Here, BL refers to the \% accuracy of baseline software CNN model and DC refers to the \% accuracy of CNN on DeepCAM.}
\label{fig:accuracy}
\vspace{-5mm}
\end{figure*}

From Fig. 2, we determined that the error in the approximate dot-product operation depends on the hash length (k). We find that each CNN layer requires a certain minimum hash length to maintain the overall classification accuracy (referred to as optimal hash length). Some layers are sensitive to a smaller hash length, while others are very robust. One way to maintain the classification accuracy would be to choose the maximum value out of all optimal hash lengths as the fixed hash length across all CNN layers. However, this would lead to hardware over-utilization.  
As a result, we propose a variable hash length encoding strategy (\textit{i.e.}, using different hash lengths corresponding to each CNN layer) that can help maintain the CNN classification accuracy, as shown in Fig.~\ref{fig:accuracy}. 
In order to have variable hash lengths, the size of the CAM module should also be varied. To this end, we propose a dynamic size CAM-based accelerator, described in the next section.

\subsection{Dynamic size CAM-based Accelerator and Dot-product Computations}
\label{sec:dynamiccam}

\begin{figure}[!t]
\centering
\includegraphics[width=0.8\linewidth]{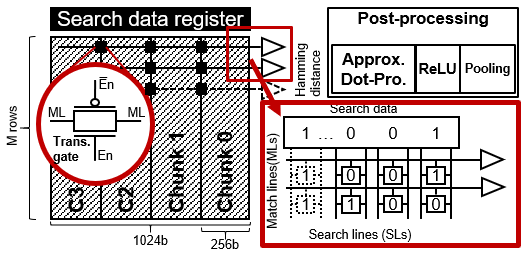}
\vspace{-5mm}
\caption{Dynamic size CAM-based accelerator for estimating the hamming distance between activations and weights in parallel before conducting the final approximate dot-product in post-processing module.}
\label{fig:dynamicsizetcam}
\vspace{-5mm}
\end{figure}

As shown in Fig~\ref{fig:dynamicsizetcam}, our dynamic size DeepCAM accelerator comprises of four chunks; the word size for each chunk is 256-bits. Each chunk is connected to its adjacent chunks by using transmission gates. The maximum word length for the CAM module can be expanded to 1024-bits. In this design, we use transmission gates (behaving as switches driven by an enable signal $En$) since the combination of both NMOS \& PMOS transistors prevent signal degradation and forward all the voltages on the bit-line to the next chunk. The sense amplifier~\cite{ni2019ferroelectric} detects the pull-down time of ML to 0-voltage and specifies the clock cycle time required for ML to attain 0-voltage, where the hamming distance between search data and row CAM data is computed. By enabling/disabling the transmission gates, we can dynamically change the word length (and hence, the hash length) from 256 to 1024 bits in the CAM module. With this flexibility for choosing the optimal hash lengths for each CNN layer during dot-product computations, our CAM can achieve lower access power as well as better energy efficiency. 

In the DeepCAM accelerator, the CAM module helps compute the hamming distances in parallel for multiple input activations and weight kernels simultaneously. Note, the computation of hamming distances (needed for approximate dot-products) occurs in CAMs with O(1) time-complexity, and this manifests as significant reduction in computation time per inference compared to Eyeriss \cite{chen2016eyeriss} as we will see in section \ref{sec:evaluation}. However, two further steps are needed to complete the approximate dot-product operations as shown in equation~\ref{equ:appdotproduct}: 1) calculating the  output of the cosine function, and 2) multiplication of the cosine output with the L2 norms of the operands. Implementing cosine functions on hardware can have a significant overhead as multiple computation cycles or lookup-tables with large memory sizes are required to calculate the cosine output using the hamming distance results from the CAM \cite{cham1989development}. To minimize the hardware costs, we apply the approximate cosine function as follows:
\begin{equation}
cosine(\theta) = \left\{
  \begin{array}{ll}
    -0.96\theta+1.51 & \frac{\pi}{3} < \theta \leq \frac{\pi}{2}\\
    1-\frac{\theta}{\pi} & 0 < \theta \leq \frac{\pi}{3}\\
   -cosine(\pi-\theta) & \theta > \frac{\pi}{2}\\
  \end{array}
\right.
\label{equ:cosine}
\end{equation}

\begin{figure}[!t]
\centering
\includegraphics[width=\linewidth]{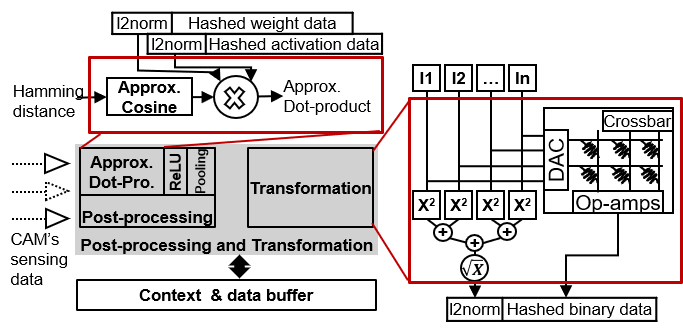}
\caption{The post-processing and transformation module includes- 1) a post-processing sub-module and 2) an online activation context generator sub-module}
\label{fig:onlinecontextgenerator}
\vspace{-5mm}
\end{figure}

After obtaining the cosine output, it is multiplied with the L2 norm of the CNN weights and activations to generate the final approximate dot-product. Our deep learning accelerator also supports peripheral operations such as ReLU, pooling, batchnorm in the digital domain that are carried out in the Post-processing \& transformation module as shown in Fig~\ref{fig:onlinecontextgenerator}.


\subsection{Activation post-processing and transformation}
\label{sec:postprocessing}

The output activations of a CNN layer, generated as a result of approximate dot-product computations, are required to be converted into activation contexts for computation in the subsequent layer. As shown in Fig.~\ref{fig:deepcam}, we could send the intermediate activations back to the software to generate the activation contexts. However, it will lead to significant energy and latency overhead owing to the data communication. Hence, we propose the on-the-fly activation context generator in hardware (part of the Post-processing \& transformation module) for converting the intermediate activations into activation contexts for the next CNN layer. Similar to the software context generator, the activation context generator will generate L2 norm and hashed binary data from the input activations. The L2 norm functionality is implemented using a simple adder tree and a digital square-root module. Further, we use a non-volatile memory (NVM) based crossbar-array to encode the random vector C (see section \ref{sec:contextgen}) as synaptic weights and implement the on-chip hash function. 
Since we only need the sign bits for carrying out projection operation using the crossbar-array, we replace the high-resolution ADCs with simple sense amplifiers that detect the negative results. This transformation module is implemented as shown in Fig.~\ref{fig:onlinecontextgenerator}.

\section{Evaluations and Results}
\label{sec:evaluation}
In this section, we will evaluate our DeepCAM accelerator using state-of-the-art pre-trained CNN models (LeNet5, VGG11, VGG16 and ResNet18) with benchmark datasets (MNIST, CIFAR10 and CIFAR100)~\cite{alzubaidi2021review}. The details are summarized in Table.~\ref{tab:evalsetup}. Note, our FeFET CAM uses variable hash length encoding strategy to maintain the inference accuracy of the CNN models on hardware close to the software accuracy as shown in Fig. \ref{fig:accuracy}. We compare our work against other deep learning hardware accelerators that are widely used for CNN inference.

\subsection{Methodology}

\begin{figure}[!t]
\centering
\includegraphics[width=\linewidth]{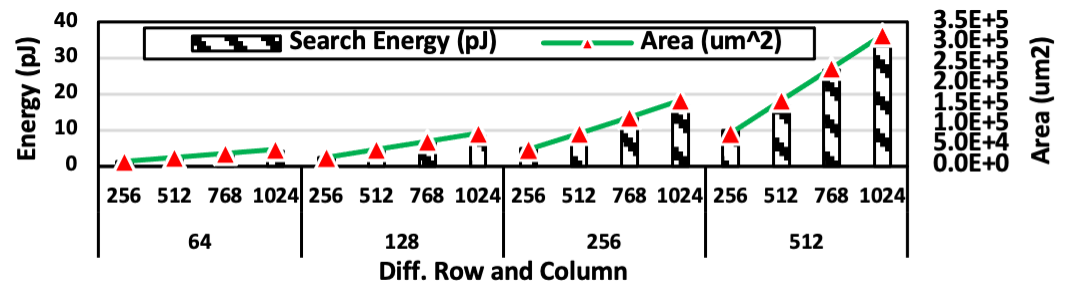}
\vspace{-10mm}
\caption{Plot of CAM-based hardware overhead results with various row and column sizes.}
\label{fig:tcameval}
\vspace{-5mm}
\end{figure}

To evaluate our DeepCAM accelerator, we carry out system-level and hardware-level simulations. For the system-level evaluation, we consider two dataflows: 1) weight-stationary, where the CAM module stores weight contexts as CAM data and activation contexts are passed as search data; 2) activation stationary, where the CAM module stores activation contexts as CAM data and weight contexts are passed as search data. The DeepCAM simulation system is implemented in the manner shown in Fig.~\ref{fig:deepcam}. For the dynamic size CAM, we can have CAM row sizes of 64/128/256/512 to store the fetched weight/activation contexts as CAM data, and CAM column sizes of 256/512/768/1024 to represent the variable context hash lengths. We use an FeFET CAM to evaluate our proposed accelerator. The FeFET CAM search energy and area statistics are extracted from EvaCAM~\cite{liu2022eva} to project the hardware overhead results for our chosen row/column sizes (see Fig~\ref{fig:tcameval}). For the hardware evaluation using DeepCAM, we implement the hardware description code and use Synopsys Design Compiler and PrimeTime~\cite{kurup2012logic,walia2009primetime} to extract the power consumption, area, and timing results. The hardware evaluations are carried out at a clock frequency of 300 MHz using 45 nm CMOS technology node. Further, we simulate the crossbar-array in the Post-processing \& transformation module having FeFET devices as synapses using the NeuroSim tool \cite{peng2019dnn+}. Both the system and hardware evaluation data are used to estimate the overall computation time and energy savings with our DeepCAM accelerator for various CNN models.  

\textbf{Baselines:} We compare our work with the state-of-the-art Eyeriss accelerator based on systolic array architecture \cite{chen2016eyeriss}. For systolic array evaluations, we modify the SCALE-Sim~\cite{samajdar2018scale} framework with appropriate network topology and systolic array configuration of Eyeriss~\cite{chen2016eyeriss}. Although, INT16 is used in ~\cite{chen2016eyeriss} as the data precision, we choose INT8 representation because INT8 is the state-of-the-art quantization for various CNN workloads \cite{jouppi2021ten}. Hence, we implement Eyeriss with a processing-array configuration of 14$\times$12 and a datapath with INT8 representation. After running SCALE-Sim on the various CNN models, we extract the  computational cycles (indicating overall computation time) and hardware utilization during inference. As a second baseline, we use Intel Skylake CPU with the AVX-512 extension that supports the vector neural network instruction ~\cite{doweck2017inside}.

\begin{table}[]
\caption{Table showing our hardware evaluation setup. Note, VHL stands for variable hash length.}
\label{tab:evalsetup}
 \centering
 \resizebox{0.8\linewidth}{!}{%
\begin{tabular}{|c|c|c|c|}
\hline
Category  & CPU & Systolic & DeepCAM \\ \hline
Configuration & \makecell{ Skylake with\\ AVX-512~\cite{doweck2017inside} }& \makecell{Eyeriss \\($14\times12$)~\cite{chen2016eyeriss}} & \makecell{FeFET CAM\\with VHL} \\ \hline
\makecell{Hardware \\performance} & \multicolumn{3}{c|}{Overall inference computation cycles}\\ \hline
\makecell{Energy \\consumption} & \multicolumn{3}{c|}{Dynamic inference energy}\\ \hline
\makecell{CNN \& \\Dataset} & \multicolumn{3}{c|}{\makecell{LeNet5\_MNIST, VGG11\_CIFAR10,\\ VGG16\_CIFAR100, ResNet18\_CIFAR100}}\\ \hline
\end{tabular}
}
\end{table}

\subsection{Hardware Performance with DeepCAM}

We find that the activation-stationary dataflow results in a lower number of computational cycles compared to the weight-stationary dataflow with multiple CNN topologies on our DeepCAM accelerator. To understand this, let us consider the following example. Suppose, we have a single-channeled input of size 32x32 and 6 weight-kernels of size 5x5 for convolution with stride 1. Then, for obtaining the output feature map, we need (28*28=784) input vectors for the 6 kernel-vectors. If weight-stationary mode of mapping is considered for a CAM with 64 rows, whereby only the 6 rows corresponding to the 6 kernels are occupied out of the 64 CAM rows, we have an hardware utilization of $6/64 = 9.4\%$. On the other hand with activation-stationary mode of mapping on the 64 CAM rows, the hardware utilization becomes $100\%$. Hence, activation-stationary mode of dataflow in DeepCAM induces full utilization of the available CAM hardware and thus, facilitates faster convolutions with an overall lower number of computational cycles. 

\begin{figure}[!t]
\centering
\includegraphics[width=\linewidth]{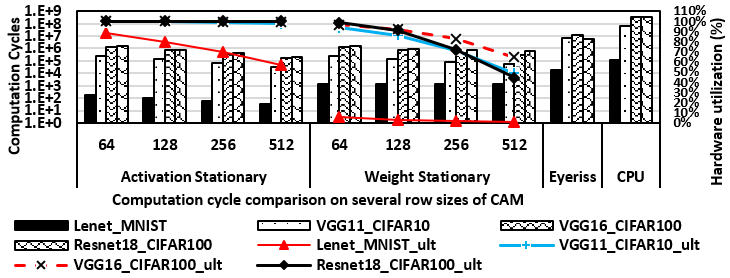}
\vspace{-8mm}
\caption{Plot of computational cycles and hardware utilization for weight/activation-stationary modes of DeepCAM compared with Eyeriss and traditional CPU.}
\label{fig:performance}
\vspace{-5mm}
\end{figure}

Compared to Eyeriss, our DeepCAM (with 64 CAM rows and activation-stationary dataflow) is $\sim523.5\times$ efficient in reducing inference computational cycles for the LeNet\_MNIST topology and $\sim3.3\times$ efficient in case of ResNet18\_CIFAR100 topology. The efficiency in reducing computational cycles increases to $\sim26.4\times$ for ResNet18\_CIFAR100 topology when CAM row size is increased to 512. Compared to Intel Skylake, our DeepCAM (with 64 CAM rows) is $\sim235.4\times$ efficient in reducing computational cycles for LeNet\_MNIST with weight-stationary dataflow and up to $\sim3498\times$ with activation-stationary dataflow. The summary of the above results is presented in Fig~\ref{fig:performance}. 

\subsection{Energy Consumption per Inference}

\begin{figure}[!t]
\centering
\includegraphics[width=\linewidth]{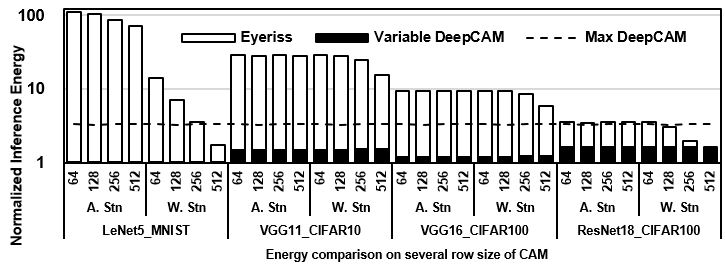}
\vspace{-9mm}
\caption{Plot of normalized energy consumption of DeepCAM compared to Eyeriss.}
\label{fig:energy}
\vspace{-5mm}
\end{figure}

In this section, we only make a comparison between our DeepCAM accelerator and Eyeriss, because traditional CPUs are known to be very energy-hungry architectures. Fig.~\ref{fig:energy} compares the energy results between our DeepCAM with variable hash lengths to that of Eyeriss. In our comparison, we choose the baseline as CNNs implemented on DeepCAM with homogeneous 256-bit hash lengths across all layers.  All results in Fig.~\ref{fig:energy} are normalized to this baseline. Also, Max DeepCAM refers to a homogeneous 1024-bit hash length implementation across layers. The variable hash length DeepCAM yields 1.78$\times$ energy reduction compared to Eyeriss in the case of LeNet\_MNIST with 512 CAM rows (in weight-stationary mode). However, we can increase the energy reduction up to 109.4$\times$ by changing the dataflow to activation stationary. In case of ResNet18\_CIFAR100, DeepCAM with variable hash length achieves energy reduction of 2.16$\times$ compared to Eyeriss. 

\section{Related works and Comparison}
\label{sec:relatedworks}
\vspace{-2mm}
\begin{table}[h!]
    \centering
    \caption{Comparison of DeepCAM (with VHL) with previous PIM works.}
    \vspace{-2mm}
    \label{tab:compare_works2}
    \resizebox{0.9\columnwidth}{!}{
    \begin{tabular}{|l|c|c|c|}
    \hline
         \multirow{1}{*}{Work} &  NeuroSim \cite{peng2019dnn+} & Valavi et al. \cite{valavi201964} & \textbf{Ours}\\ \hline
         Device & RRAM & SRAM & \textbf{FeFET} \\ \hline
         Dot-Product Mode & Algebraic & Algebraic & \textbf{Geometric} \\
         \hline
         
        Energy per & \multirow{2}{*}{34.98} & \multirow{2}{*}{3.55} & \multirow{2}{*}{\textbf{0.488}} \\
        Inference ($\mu$J) & & & \\
        \hline
        Computation Cycles & \multirow{2}{*}{5.74} & \multirow{2}{*}{2.56} & \multirow{2}{*}{\textbf{2.652}} \\
        per Inference ($\times 10^5$)& & & \\
        \hline
    \end{tabular}}
\vspace{-5mm}
\end{table}
Utilizing CAM architectures for deep learning applications is a relatively new research direction. Prior works such as  \cite{ni2019ferroelectric, reis2020fast, laguna2021memory,lagunahardware, li2022imars} have used CAMs as associative memories for fast and energy-efficient search operations across various deep learning workloads. In \cite{ni2019ferroelectric}, the classifier layer of a DNN is implemented using FeFET CAMs operating on Locality-sensitive Hashing (LSH). However, compared to other DNN layers, the classifier has much lower computational overhead. Thus, the application of CAM to implement DNN classifier does not speed up the deep learning system and also incurs additional power consumption by the CAM array. Another work~\cite{reis2020fast} uses Range-encoding (RE) method for data storage to perform few-shot learning tasks. However, the proposed design requires a significant number of CAM accesses to measure the $L_{\infty}$ and $L_{1}$ distance and hence, is computationally intensive. We know that dot-product operations are the key computational kernels for deep learning. However, developing large-scale CAM-based deep learning accelerators has been challenging because transforming CAMs from being associative memories to efficient dot-product engines has not been well explored. To this end, exploiting the properties of CAM to estimate the hamming-distance between input activations and weights (using random projection hashing method) in a DNN and performing energy-efficient approximate geometric dot-products have been the key contributions of our work. We have shown that our DeepCAM accelerator opens up possibilities to carry out highly parallelized dot-product operations on hardware for large-scale deep learning tasks. 

Now, we compare our PIM-based DeepCAM architecture with two previously proposed PIM-based works \cite{peng2019dnn+, valavi201964} for the acceleration of deep learning workloads. Both of these works conduct inference of CNNs on analog compute macros (based on SRAM or NVM devices) by computing algebraic dot-products. Analog dot-product PIM engines have been shown to facilitate compact and energy-efficient implementation of DNNs on hardware with high parallelism \cite{chakraborty2020pathways}. Table.~\ref{tab:compare_works2} compares DeepCAM against a VGG11\_CIFAR10 CNN evaluated on RRAM device-based PIM engine using the NeuroSim tool \cite{peng2019dnn+} and SRAM-based PIM engine as described in \cite{valavi201964}, in terms of dynamic energy and computation cycles per inference. We find that our PIM-based solution (DeepCAM with variable hash length) is $\sim71.68 \times$ more energy-efficient and requires $\sim2.16 \times$ less computation cycles per inference than \cite{peng2019dnn+}. In comparison to \cite{valavi201964}, DeepCAM is $\sim7.27 \times$ energy-efficient, but requires slightly higher computational cycles per inference.

\section{Conclusion}
\label{sec:conlcusion}
This paper proposes DeepCAM, a reconfigurable CAM-based inference accelerator built on critical innovations to alleviate computation time demands of deep learning workloads. We find that DeepCAM can be up to $523\times$ faster than Eyeriss -conventional systolic array architecture while consuming up to $109\times$ less energy than Eyeriss. All these savings come with negligible loss in output quality in image recognition tasks. 

\section*{Acknowledgement}
\small
This work was supported in part by C-BRIC, a JUMP center sponsored by DARPA and SRC, Google Research Scholar Award, the National Science Foundation (Grant \#1947826), TII (Abu Dhabi), the DARPA AI Exploration (AIE) program, and the DoE MMICC center SEA-CROGS (Award \#DE-SC0023198).
\normalsize


\bibliographystyle{IEEEtranS}
\bibliography{refs}

\end{document}